\documentclass[journal]{IEEEtran}
\usepackage[T1]{fontenc}
\usepackage{graphicx}
\usepackage{times}
\usepackage{helvet}
\usepackage{courier}
\usepackage{amsmath}
\usepackage{algorithm}
\usepackage{algorithmic}
\usepackage{csquotes} 
\usepackage{color}
\usepackage{paralist}
\usepackage{amssymb}
\usepackage{indentfirst}
\usepackage{subfigure}
\usepackage{float}
\usepackage{multirow}
\usepackage{cite}
\usepackage{mathrsfs}

\usepackage{mathrsfs} 
\usepackage{amsfonts}
\usepackage{todonotes}
\usepackage{pgfplots} 
\usepackage{epstopdf}
\usepackage{epsfig}
\pgfplotsset{compat=newest}
\usepackage{color}
\definecolor{forestgreen}{RGB}{0,139,69}

\usepackage{xcolor}
\definecolor{citecolor}{HTML}{0071bc}
\usepackage[colorlinks, linkcolor=red,  anchorcolor=blue, citecolor=citecolor]{hyperref} 

\usepackage{xcolor}
\definecolor{SeaGreen4}{RGB}{0,205,102} 
\definecolor{SlateBlue}{RGB}{106,90,205} 
\definecolor{DarkRed}{RGB}{178,34,34} 
	
\usepackage[switch]{lineno}

\usepackage{textcomp,booktabs}
\usepackage{amssymb}
\usepackage{pifont}

\usepackage{makecell}

\usepackage{colortbl}
\definecolor{mygray}{gray}{.9}
\definecolor{mypink}{rgb}{.99,.91,.95}
\definecolor{mycyan}{cmyk}{.3,0,0,0}

\begin{document}

\title{ MGRL-RSCC: Multi-Granularity Reward Reinforcement Learning for Fine-Grained Remote Sensing Change Captioning  }   

\author{Futian Wang, Mengqi Wang, Xiao Wang*, \emph{Member, IEEE}, Wentao Wu, Haowen Wang, Zhicheng Zhao, Jin Tang 

\thanks{ $\bullet$ Futian Wang, Mengqi Wang, Xiao Wang, Haowen Wang, Jin Tang are with the School of Computer Science and Technology, Anhui University, Hefei 230601, China. (email: \{xiaowang, tangjin, wanghaowen\}@ahu.edu.cn, e24301148@stu.ahu.edu.cn)} 

\thanks{ $\bullet$ Wentao Wu, Zhicheng Zhao are with School of Artificial Intelligence, Anhui University, Hefei 230601, China. (email: wa22201027@stu.ahu.edu.cn, zhaozhicheng@ahu.edu.cn)} 

\thanks{* Corresponding Author: Xiao Wang}   
}

\markboth{ IEEE Transactions on ***, 2026 } 
{Shell \MakeLowercase{\textit{et al.}}: Bare Demo of IEEEtran.cls for IEEE Journals}

\maketitle

\begin{abstract}
Remote Sensing Change Captioning (RSCC), which aims to generate accurate and detailed linguistic descriptions of ground object variations from bi-temporal remote sensing images, is a critical and challenging task in intelligent remote sensing interpretation. The mainstream autoregressive training paradigm faces severe exposure bias and train-test distribution mismatch, resulting in cumulative generation errors. They tend to produce conservative and template-fixed captions while ignoring subtle scene change details. To address these challenges, this paper proposes a novel multi-granularity reward reinforcement learning paradigm, termed MGRL-RSCC. Specifically, we first leverage a CNN and hierarchical self-attention module to extract and enhance visual features from bi-temporal remote sensing images. A Transformer decoder is then utilized to complete visual-to-linguistic translation. Different from existing methods, we design a dual-decoding strategy and a two-stage joint optimization scheme, which combines token-level supervised learning via greedy decoding and multi-granularity reward-driven self-critical reinforcement learning via sampling decoding. We further construct three complementary reward functions covering linguistic fluency, change state consistency, and structural-semantic relevance to comprehensively optimize caption quality and alleviate false and missing change descriptions. Extensive experiments on multiple public RSCC benchmark datasets demonstrate that the proposed MGRL-RSCC effectively mitigates exposure bias and conservative generation problems in traditional autoregressive methods. 
The source code and pre-trained models will be released on 
\url{https://github.com/Event-AHU/MGRL-RSCC}. 
\end{abstract}


\IEEEpeerreviewmaketitle

\section{Introduction}

\IEEEPARstart{R}{emote} Sensing Change Captioning (RSCC)~\cite{hoxha2022change,liu2022remote} aims to accurately identify changes in the location, morphology, category and attributes of ground objects by comparatively analyzing remote sensing images of the same area acquired at different times. Meanwhile, it excavates detailed features, change types, and evolution laws of changed regions. As one of the core fundamental tasks for intelligent interpretation of remote sensing imagery, this task possesses extremely high practical application and scientific research value in numerous critical fields including territorial spatial planning, dynamic monitoring of natural resources, emergency disaster assessment, urban construction renewal, and ecological environment governance. It serves as a core technique supporting intelligent and refined applications of Earth observation. In recent years, a wide variety of intelligent algorithms~\cite{zou2025remote} have emerged continuously, greatly improving the efficiency and automation of change information extraction from remote sensing images. Nevertheless, RSCC has not yet been thoroughly and perfectly addressed, and existing algorithm systems suffer from prominent inherent limitations. Therefore, continuous in-depth research on high-performance and robust remote sensing change captioning is of great significance for improving the refinement level of intelligent remote sensing image interpretation and expanding the practical application scenarios of remote sensing technology.

\begin{figure*}
\centering
\includegraphics[width=1\textwidth]{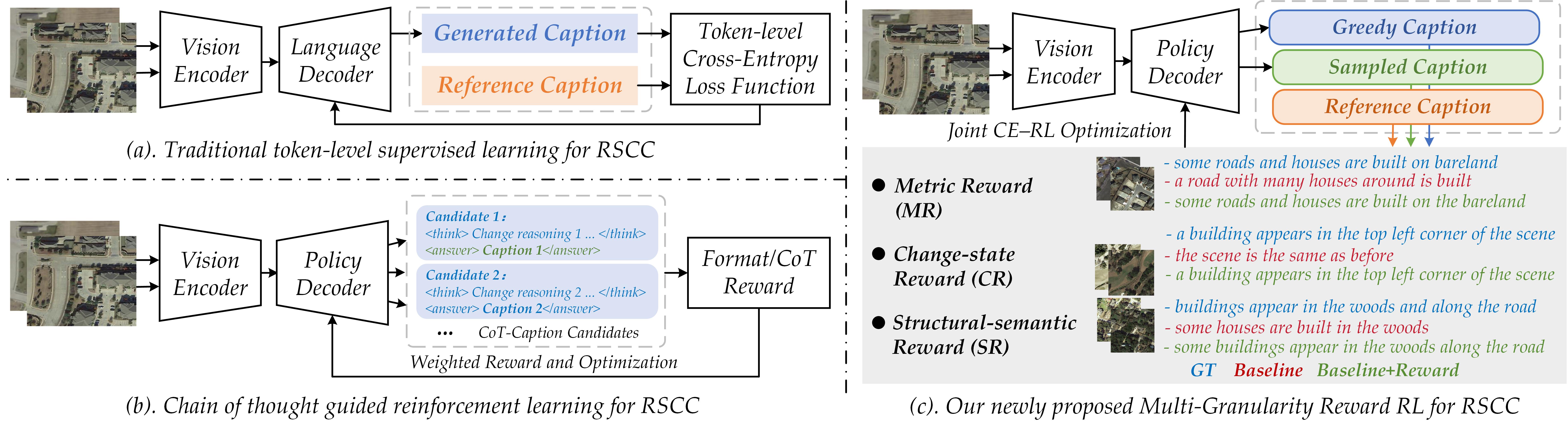}
\caption{Comparison between the existing RSCC framework and our newly proposed MGRL-RSCC.} 
\label{fig:firstIMG}
\end{figure*}

According to our observation, remote sensing change captioning has roughly undergone the following evolutionary stages: CNN-based methods~\cite{hoxha2022change}, autoregressive generation paradigms built upon Transformers~\cite{liu2022remote, chang2023changes, liu2023progressive}, large foundation models~\cite{zhu2024semantic, yang2025enhancing, deng2025changechat}, and reinforcement learning-based paradigms~\cite{zhang2026rsc}. Specifically, Fig.~\ref{fig:firstIMG} (a) illustrates the conventional token-level supervised paradigm, where the visual encoder and language decoder are jointly optimized using cross-entropy loss. Representative methods include PSNet~\cite{liu2023progressive}, which captures multi-scale bi-temporal differences, RSCaMa~\cite{liu2024rscama}, which enables efficient feature interaction through state-space modeling, and MModalCC~\cite{karaca2025robust}, which improves robustness to realistic image degradations.
In contrast, Fig.~\ref{fig:firstIMG} (b) presents a chain-of-thought-guided reinforcement learning paradigm. RSC-CoT~\cite{zhang2026rsc} introduces visual chain-of-thought reasoning and reinforced optimization to improve interpretability and semantic consistency. Related methods such as KCFI~\cite{yang2025enhancing} and ChangeVG~\cite{xue2026towards} further enhance key-change perception and interactive change understanding.

Despite remarkable progress, we argue that existing algorithms still suffer from the following limitations: 
1). Built upon the Transformer architecture, mainstream RSCC algorithms~\cite{liu2022remote,chang2023changes,liu2023progressive} employ token-level Auto-Regressive (AR) training paradigms that suffer from exposure bias~\cite{bengio2015scheduled}, wherein the distribution mismatch between training and inference leads to cumulative errors throughout the generated sequence. 
2). Constrained by token-wise cross-entropy loss, the AR-based models~\cite{rennie2017self} exhibit conservative generation behavior, favoring high-frequency words and fixed sentence templates while neglecting subtle scene details in remote sensing change description. 
3). Current reinforcement learning-based RSCC algorithms~\cite{zhang2026rsc} integrate chain-of-thought reasoning~\cite{wei2022chain} into the change description pipeline, enhancing model interpretability. Nevertheless, this approach substantially increases training and inference costs, restricting model deployment on low-computational-power platforms. 
Therefore, it is natural to raise the following question: ``\textit{How can we devise a novel reinforcement learning-augmented autoregressive paradigm for RSCC to alleviate exposure bias, mitigate conservative generation, and attain competitive performance without incurring the substantial computational overhead of explicit chain-of-thought reasoning?}"

To address these issues, in this paper, we propose a novel \textbf{M}ulti-\textbf{G}ranularity \textbf{R}eward reinforcement \textbf{L}earning paradigm for \textbf{R}emote \textbf{S}ensing \textbf{C}hange \textbf{C}aptioning, termed \textit{MGRL-RSCC}. 
As shown in Fig.~\ref{fig:framework}, given bi-temporal remote sensing images, we first extract visual embeddings using a CNN (Convolutional Neural Network) and enhance the global representation with a hierarchical self-attention module. Then, a Transformer decoder is adopted to project visual tokens into language descriptions. Unlike existing methods, we employ two decoding strategies, including greedy decoding and sampling decoding, the former enables autoregressive generation to optimize token-level supervised loss, and the latter adopts multi-granularity reward-driven reinforcement learning for multi-dimensional enhancement. In more detail, three types of reward functions are considered, including linguistic metric reward, change-state reward, and structural-semantic reward. Thus, a two-stage joint optimization framework is proposed that combines token-level supervised learning and self-critical reinforcement learning for remote sensing change captioning.

To sum up, the main contributions of this paper can be summarized as follows: 

1). We propose a novel {M}ulti-{G}ranularity {R}eward reinforcement {L}earning paradigm for RSCC, termed \textit{MGRL-RSCC}, which integrates linguistic metric reward, change-state reward, and structural-semantic reward. It comprehensively evaluates generated captions from linguistic fluency, change consistency, and fine-grained semantic relevance, and alleviates the problem of false or missing change descriptions. 

2). We construct a two-stage joint optimization framework that combines \textit{token-level supervised learning} and \textit{self-critical reinforcement learning} for remote sensing change captioning, which mitigates exposure bias and the train-test mismatch in traditional autoregressive caption generation. 

3). Extensive experiments on multiple benchmark datasets (i.e., LEVIR-CC~\cite{liu2022remote}, Dubai-CC~\cite{hoxha2022change}, WHU-CDC~\cite{shi2024multi}) fully validated the effectiveness of our newly proposed multi-granularity reward function for the RL-based RSCC task. 

\textit{The rest of this paper is organized as follows:} In Section~\ref{sec::relatedWorks}, we review the related works from RSCC, RL-based sequence generation, and graph-structured semantic modeling. Then, we introduce the methodology in Section~\ref{sec::method}, with a focus on the overview, vision encoder and captioning decoder networks, multi-granularity reward function, and self-critical RL. In Section~\ref{sec::experiments}, we validate the effectiveness of MGRL-RSCC based on qualitative and quantitative experiments. We conclude this paper and discuss future work in Section~\ref{sec::conclusion}.

\section{Related Works} \label{sec::relatedWorks}

\subsection{Remote Sensing Change Captioning}
Remote sensing change captioning extends conventional change detection from pixel-level localization to sentence-level semantic description, requiring models to identify changed regions, recognize changed objects, and express temporal differences in natural language. Foundational studies such as RSICCformer~\cite{liu2022remote} and Chg2Cap~\cite{chang2023changes} established the basic paradigm of bi-temporal visual encoding and language decoding for change description. Subsequent works further improved visual interaction and change localization. ICT-Net~\cite{cai2023interactive} enhances cross-temporal feature interaction with an interactive change-aware Transformer, while SFT~\cite{sun2024lightweight} introduces sparse focus attention to emphasize salient changed regions with lower computational cost. More recent work has shifted toward broader data settings, stronger robustness, and richer interaction. SECOND-CC and MModalCC~\cite{karaca2025robust} emphasize realistic degradation factors such as illumination variation, blur, viewpoint changes, and registration errors, while enriching supervision with semantic segmentation maps. RSCC~\cite{chen2026rscc} further expands the task to disaster scenarios with a substantially larger pre-/post-event benchmark for change-aware vision-language learning. In parallel, ChangeVG~\cite{xue2026towards} extends the problem from single-output description to a more comprehensive interactive change-understanding setting, jointly covering change captioning, binary change classification, counting, and localization under instruction tuning. On the modeling side, Semantic-CC~\cite{zhu2024semantic} leverages foundational knowledge and pixel-level semantic guidance to improve fine-grained change expression, PM3Net~\cite{qu2026mask} introduces mask-guided multigranular Mamba modeling for efficient spatiotemporal representation, and RotCap~\cite{zhao2026disturbance} improves robustness under rotation disturbance through self-supervised multifrequency representation. Despite this progress, most recent methods still focus primarily on strengthening visual encoding and fusion, while the semantic faithfulness of the generated description remains only indirectly constrained during training.

\subsection{Reinforcement Learning based Sequence Generation}
Reinforcement learning has been widely adopted in sequence generation to reduce the mismatch between token-level supervision and sequence-level evaluation. Classical methods such as SCST~\cite{rennie2017self} and actor-critic training~\cite{zhang2017actor} showed that directly optimizing sentence-level rewards can alleviate exposure bias and better align training with evaluation. In remote sensing description, VRTMM~\cite{shen2020remote} combines variational representation learning with reinforcement learning to improve caption quality, while ADCM~\cite{chavhan2021novel} introduces an actor dual-critic strategy to provide more informative reward feedback for remote sensing image captioning. More recent studies have moved beyond optimizing only n-gram-based metrics and started to design richer reward signals. SC-Captioner~\cite{zhang2025sc} improves image captioning through a self-correction framework, where rewards are computed from object, attribute, and relation sets extracted by scene-graph parsing. CapRL~\cite{xing2025caprl} further reformulates caption optimization with verifiable rewards, measuring caption quality by whether a vision-free language model can answer questions using the generated caption alone. CCCaption~\cite{tang2026cccaption} separates caption quality into completeness and correctness, and explicitly optimizes these two aspects with a dual-reward reinforcement learning scheme. This recent trend is particularly relevant to remote sensing change description, where useful outputs should not only be fluent or lexically similar to references, but also preserve correct change states, object details, and semantic relations.

\subsection{Graph-Structured Semantic Modeling}
Graph-based modeling provides an effective way to represent entities, relations, and contextual dependencies beyond flat token sequences. In general vision-language tasks, SGAE~\cite{yang2019auto} encodes scene graphs to improve structural consistency, SG2Caps~\cite{nguyen2021defense} revisits scene graphs for image captioning, and KG-Transformer~\cite{zhang2021image} incorporates knowledge graph information into Transformer-based caption generation. Recent work in remote sensing image captioning has also started to revisit this direction from a more structured semantic perspective. CASK~\cite{li2024learning} learns consensus-aware semantic knowledge through concept correlations, and TextGCN-based decoding~\cite{das2024textgcn} introduces graph-structured word relations to improve remote sensing image caption generation. SFDR~\cite{liu2025semantic} integrates semantic-spatial feature fusion with dynamic graph refinement, using graph attention and dynamic weighting to strengthen object-level relevance and contextual grounding. For change-oriented description, SGD-RSCCN~\cite{sun2025scene} combines scene-graph construction with dependency grammar to improve both change understanding and sentence naturalness. In the broader image captioning literature, AKGMA~\cite{li2026knowledge} introduces adaptive knowledge-graph-guided multimodal alignment to reduce knowledge hallucination and improve open-world semantic consistency. These studies confirm that structured priors are useful for modeling objects, relations, and scene context. However, most existing methods inject graph information directly into the encoder or decoder, which tightly couples structured reasoning with feature extraction. Different from this line, our method uses graph-structured knowledge as a reward-side semantic constraint during reinforcement learning, encouraging the generated description to remain consistent with object categories, change relations, spatial contexts, and valid triplet structures.

\section{Our Proposed Approach} \label{sec::method}

\begin{figure*}
\centering
\includegraphics[width=1\textwidth]{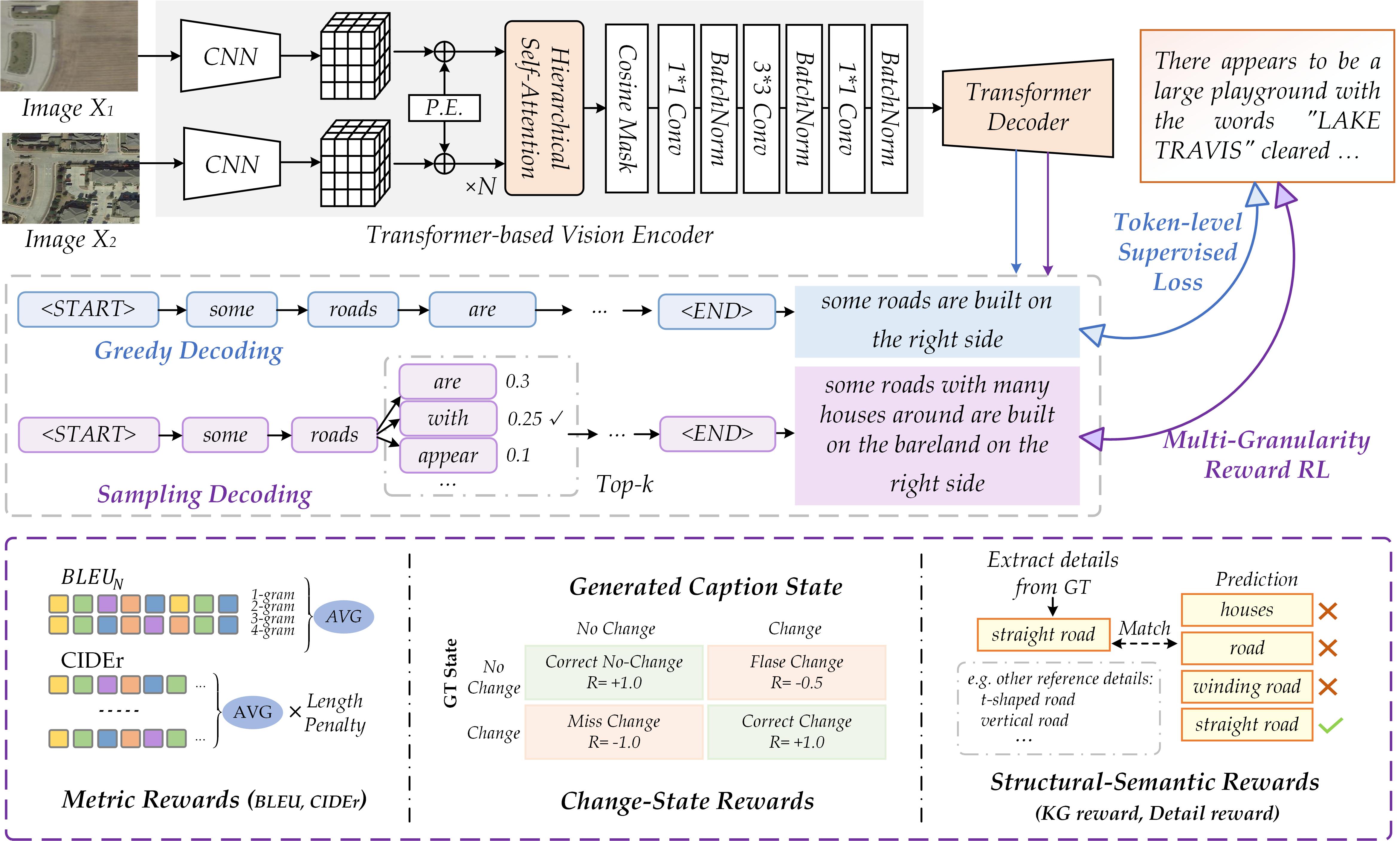}
\caption{
An overview of our proposed Multi-Granularity Reward Reinforcement Learning paradigm for RSCC, termed MGRL-RSCC, which integrates linguistic metric reward, change-state reward, and structural-semantic reward. 
}
\label{fig:framework}
\end{figure*}




\subsection{Overview}

Given a pair of bi-temporal remote sensing images, our goal is to generate a caption that accurately describes the semantic changes between the two observations. As illustrated in Fig.~\ref{fig:framework}, the proposed MGRL-RSCC framework consists of a bi-temporal visual encoder, a Transformer-based caption decoder, and a multi-granularity reward-guided reinforcement learning framework. The visual encoder extracts and enhances the pre-change and post-change features, while the decoder autoregressively generates a change description from the fused change-aware representation.

To reduce the mismatch between token-level supervised training and sequence-level caption evaluation, we employ a two-stage optimization strategy. The captioning model is first trained with teacher-forced cross-entropy learning and is subsequently fine-tuned using self-critical sequence training (SCST) while retaining the cross-entropy objective as supervised regularization. In addition to conventional caption-level metric rewards, we introduce semantic supervision at multiple granularities through three complementary reward groups comprising five reward components: caption-level metric rewards based on BLEU and CIDEr, a scene-level change-state reward, and fine-grained structural-semantic rewards consisting of a detail reward and a knowledge-graph reward. These reward groups jointly evaluate the generated caption in terms of sequence-level reference correspondence, global change-state correctness, and fine-grained object, relational, and contextual information.

\subsection{Input Encoder and Caption Decoder Network}



Let $I^{pre}$ and $I^{post}$ denote the pre-change and post-change remote sensing images, respectively. A shared convolutional encoder $E(\cdot)$ is first employed to extract their visual feature maps:
\[
F^{pre} = E(I^{pre}), \quad F^{post} = E(I^{post}).
\]
The extracted features are subsequently processed by an attention-based visual encoder to capture spatial context within each temporal observation and interactions across the two temporal observations. Positional embeddings are added to the visual tokens before attention-based feature enhancement.

A change-aware visual memory is then constructed from the enhanced bi-temporal features. Specifically, the pre-change and post-change features are concatenated, while their spatial cosine similarity is incorporated as an additional change-sensitive cue to highlight temporal discrepancies. The resulting representation is projected through a $1\times1$ convolution and further refined by a residual block before being flattened into a sequence of visual memory tokens. Conditioned on this visual memory, a Transformer decoder autoregressively generates the change caption using causal self-attention and cross-attention over the encoded visual representation.

\subsection{Multi-Granularity Reward Function} 

We define a unified reward function to evaluate a generated caption from multiple semantic levels. For a generated caption $y$ and its reference captions $\mathcal{Y}$, the total reward at epoch $t$ is formulated as
\[
R(y, \mathcal{Y}, t) =
\frac{
\sum_{m} \alpha_m(t) \lambda_m R_m(y, \mathcal{Y})
}{
\sum_{m} \alpha_m(t) \lambda_m
},
\]
where $R_m$ denotes an individual reward component, $\lambda_m$ is its weight, and $\alpha_m(t)$ indicates whether this reward is activated at epoch $t$. This design allows different rewards to be introduced at different training stages.



\noindent $\bullet$ \textbf{Caption-level metric rewards.} 
The first group consists of BLEU and CIDEr rewards, which measure the sequence-level correspondence between a generated caption and its reference captions. Given a generated token sequence $y$ and a set of reference captions $\mathcal{Y}=\{y_1^*,\ldots,y_K^*\}$, we first remove special tokens, including \texttt{<START>}, \texttt{<END>}, and \texttt{<NULL>}, and convert the remaining token indices into sentences. The generated caption is treated as a single hypothesis, while all available reference captions are jointly used as its comparison set.

BLEU evaluates the local lexical correspondence between the generated caption and the references using different orders of $n$-gram overlap. Instead of using only BLEU-4, we compute four sentence-level BLEU scores and average them to obtain the BLEU reward:
\[
R_{\mathrm{BLEU}}(y,\mathcal{Y})
=
\frac{1}{4}
\sum_{n=1}^{4}
\mathrm{BLEU}\text{-}n(y,\mathcal{Y}).
\]
This formulation simultaneously considers individual words, short phrases, and longer local expression patterns. CIDEr evaluates caption consensus using TF-IDF-weighted $n$-grams:
\[
R_{\mathrm{CIDEr}}(y,\mathcal{Y})
=
\mathrm{CIDEr}(y,\mathcal{Y}).
\]
It assigns larger weights to informative expressions that are shared by the reference captions and reduces the influence of frequently occurring but less discriminative words. Both rewards are computed independently for each generated caption. BLEU provides direct supervision for local lexical and phrase-level correspondence, whereas CIDEr measures the overall consensus between the generated caption and multiple human descriptions. These complementary signals guide sequence-level optimization toward captions that better match the linguistic patterns and informative expressions in the reference descriptions.

\begin{figure}
    \centering
    \includegraphics[width=0.85\linewidth]{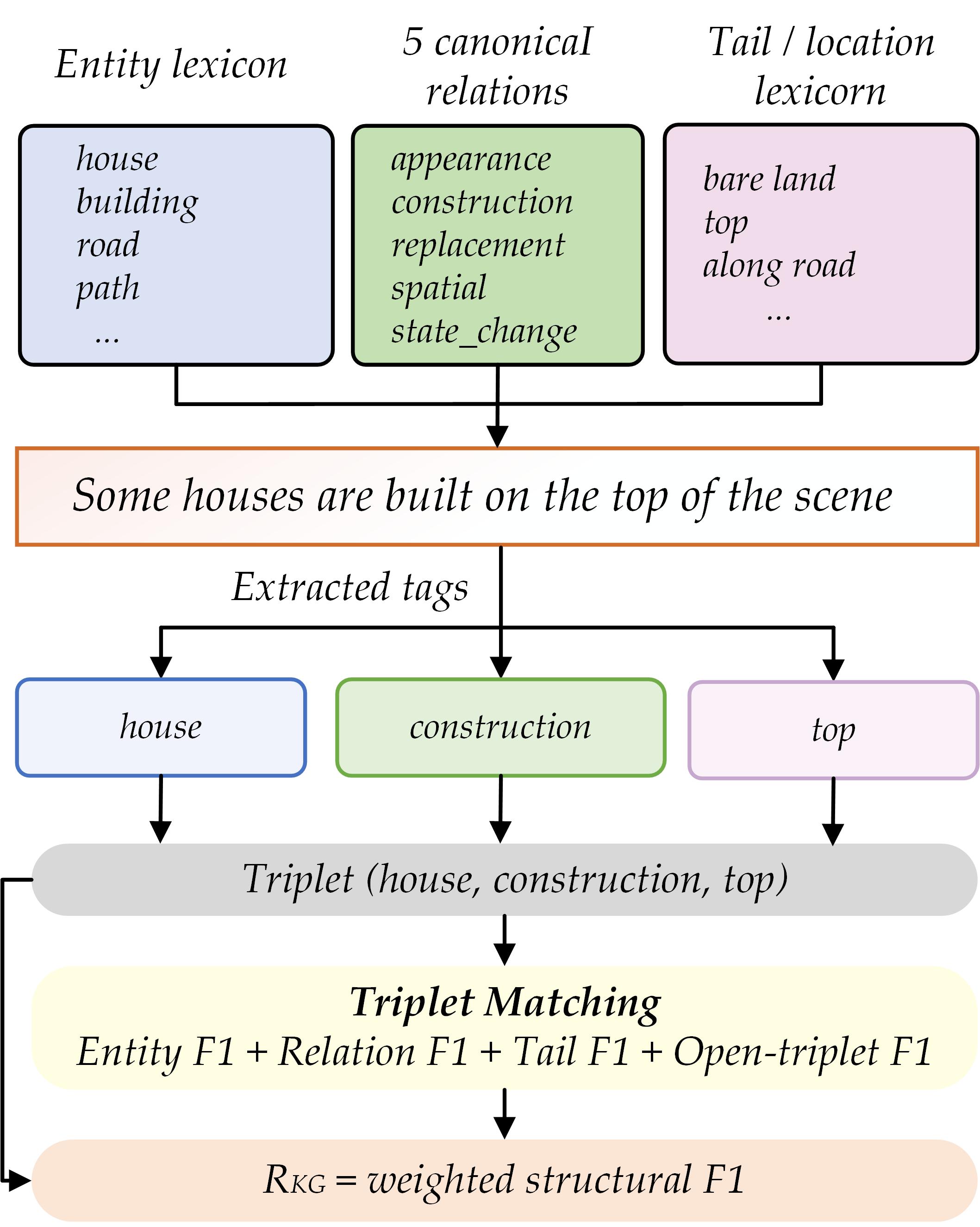}
    \caption{Illustration of the knowledge-graph reward computation, including semantic tag extraction, open-triplet construction, and triplet matching.}
    \label{fig:kg_reward}
\end{figure}

\noindent $\bullet$ \textbf{Scene-level change-state reward.}
The second group contains the change-state reward, which determines whether the generated caption correctly represents the global change status of the bi-temporal image pair. 
Unlike the metric rewards, this component does not directly evaluate lexical overlap or fine-grained object-level semantics.
Instead, it explicitly distinguishes between changed and unchanged scenes, providing a global semantic constraint for caption generation.

We define a normalized set of no-change expressions, including ``there is no difference,'' ``the scene is the same as before,'' ``the two scenes seem identical,'' and ``no change has occurred.'' Before matching, each caption is converted to lowercase, leading and trailing spaces are removed, and punctuation marks are discarded. The reference image pair is classified as a no-change scene only when all five reference captions are recognized as normalized no-change expressions. Otherwise, the reference pair is regarded as a changed scene. The generated caption is classified using the same normalization and expression-matching procedure.

Let $c(\mathcal{Y})$ and $c(y)$ denote the reference and predicted change states, respectively, where 0 represents no change and 1 represents change. The change-state reward is defined as
\[
R_{\mathrm{state}}(y,\mathcal{Y})
=
\begin{cases}
1.0, & c(y)=c(\mathcal{Y}),\\
-0.5, & c(\mathcal{Y})=0 \ \text{and}\ c(y)=1,\\
-1.0, & c(\mathcal{Y})=1 \ \text{and}\ c(y)=0.
\end{cases}
\]
A generated caption receives a positive reward when its global change state agrees with that of the references. If an unchanged image pair is incorrectly described as changed, the caption receives a penalty of $-0.5$. If a genuinely changed image pair is incorrectly described as unchanged, a larger penalty of $-1.0$ is assigned. We assign a larger penalty to missed changes than to false change predictions. The asymmetric penalty places greater emphasis on avoiding missed changes. This reward is intentionally restricted to global change-state correctness, while object categories, spatial details, and structural relations are evaluated by the fine-grained structural-semantic rewards introduced subsequently.

\noindent $\bullet$ \textbf{Fine-grained structural-semantic rewards.}
The third group consists of the detail reward and the knowledge-graph reward. The detail reward focuses on fine-grained change information that may not be sufficiently reflected by general captioning metrics, such as specific road types, numbers of houses, object attributes, and local spatial details. We construct a predefined detail lexicon containing phrases such as ``new road,'' ``t-shaped road,'' ``winding road,'' ``small house,'' and ``row of houses.'' After removing special tokens and converting words to lowercase, we scan the generated and reference captions using contiguous $n$-grams of different lengths. The matched phrases are collected as sets to avoid repeatedly counting the same detail.

For each image pair, detail phrases extracted from its five reference captions are aggregated to form the reference detail set $D(\mathcal{Y})$. The generated caption is processed using the same phrase-matching procedure to obtain $D(y)$. The detail reward is then computed as
\[
R_{detail}(y,\mathcal{Y})
=
\frac{|D(y)\cap D(\mathcal{Y})|}
{|D(\mathcal{Y})|}.
\]
This reward measures how many reference details are preserved in the generated caption. If no predefined detail phrase is detected in the reference captions, the reward is set to zero to avoid introducing an invalid supervision signal.

The knowledge-graph reward further evaluates whether the generated caption preserves structured semantic information consistent with the reference captions. After token normalization, each caption is mapped into entity tags, relation tags, and tail/context tags using phrase-level semantic lexicons. Entity tags represent remote sensing objects such as buildings, houses, roads, vegetation, bare land, and water. Relation tags describe semantic relations and change actions, including appearance, construction, replacement, spatial association, and state change. Tail/context tags represent object contexts and spatial regions, such as left, right, center, corners, and road-related locations. Different linguistic expressions referring to the same semantic concept are mapped to a shared canonical tag, thereby reducing the influence of lexical variation. An overview of the knowledge-graph reward computation is shown in Fig.~\ref{fig:kg_reward}.

Based on the detected tag spans, candidate triplets are extracted in the form $(h,r,t)$, where $h$, $r$, and $t$ denote the head entity, relation, and tail/context, respectively. Each detected relation span is treated as an anchor. We select the nearest entity span on its left as the head and the nearest context spans on its right as candidate tails. If no candidate exists in the preferred direction, the nearest available span is used instead. The resulting triplets explicitly represent the object-relation-context structures described in the generated and reference captions.

For each generated caption and reference caption, we compute entity F1, relation F1, tail/context F1, and open-triplet F1. The tag-level F1 scores are calculated from the precision and recall of the corresponding predicted and reference tag sets. For triplet matching, the similarity between two triplets is determined by the proportion of matched head, relation, and tail elements. Based on this similarity, soft precision and recall are computed between the predicted and reference triplet sets, and their harmonic mean is used as the open-triplet F1 score. This soft matching strategy provides partial credit when a predicted triplet contains some correct semantic elements but does not completely match the reference triplet.

Since an image pair is associated with multiple reference captions, the structural score is computed against each reference independently, and the highest score is selected. The final knowledge-graph reward is formulated as: 
\begin{equation}
R_{KG}(y, \mathcal{Y}) = \max_{y^*\in\mathcal{Y}}[w_e F1_e + w_r F1_r + w_t F1_t + w_o F1_{open}],
\end{equation}
where $w_e$, $w_r$, $w_t$, and $w_o$ denote the weights of the entity, relation, tail/context, and open-triplet scores, respectively. By jointly evaluating semantic elements and their triplet structures, this reward encourages the generated caption to remain consistent with the object-relation-context information expressed in the references, rather than relying only on word-level similarity.

\subsection{Self-critical Reinforcement Learning}

The model is first trained with the standard cross-entropy loss:
\[
\mathcal{L}_{CE}
=
-\sum_{i=1}^{T}
\log p_{\theta}(y_i^* \mid y_{<i}^*, I^{pre}, I^{post}).
\]
After the reinforcement learning stage starts, the decoder generates two captions for each image pair: a greedy caption $y^g$ and a sampled caption $y^s$. The greedy caption is used as the self-critical baseline, and the sampled caption is used for policy optimization.

The advantage is defined as the reward difference between the sampled caption and the greedy caption:
\[
A = R(y^s, \mathcal{Y}) - R(y^g, \mathcal{Y}).
\]
The reinforcement learning loss is then computed as
\[
\mathcal{L}_{RL}
=
- A
\sum_{i=1}^{T_s}
\log p_{\theta}(y_i^s \mid y_{<i}^s, I^{pre}, I^{post}).
\]
Finally, we combine cross-entropy learning and reinforcement learning:
\[
\mathcal{L}
=
\lambda \mathcal{L}_{CE}
+
(1-\lambda)\mathcal{L}_{RL}.
\]
This mixed objective preserves the stability of supervised training while allowing the model to directly optimize the proposed multi-granularity sequence-level rewards.






\begin{table*}
\centering
\caption{Comparisons with state-of-the-art RSCC methods on the LEVIR-CC dataset. The best and second-best results are indicated in bold and underlined, respectively.}
\label{tab:comparison LEVIR-CC}
\resizebox{0.95\textwidth}{!}{
\begin{tabular}{l|l|ccccccc}
\hline
Method & Publication & BLEU-1 & BLEU-2 & BLEU-3 & BLEU-4 & METEOR & ROUGE-L & CIDEr \\
\hline
Capt-Rep-Diff~\cite{park2019robust} & ICCV 2019 & 72.90 & 61.98 & 53.62 & 47.41 & 34.47 & 65.64 & 110.57 \\
Capt-Att~\cite{park2019robust} & ICCV 2019 & 77.64 & 67.40 & 59.24 & 53.15 & 36.58 & 69.73 & 121.22 \\
Capt-Dual-Att~\cite{park2019robust} & ICCV 2019 & 79.51 & 67.23 & 57.46 & 36.56 & 37.16 & 69.19 & 124.42 \\
DUDA~\cite{park2019robust} & ICCV 2019 & 81.44 & 72.22 & 64.67 & 57.79 & 37.15 & 71.04 & 124.32 \\
MCCFormer-S~\cite{qiu2021describing} & ICCV 2021 & 79.90 & 70.26 & 62.68 & 56.36 & 39.60 & 69.46 & 120.39  \\
MCCFormer-D~\cite{qiu2021describing} & ICCV 2021 & 80.42 & 70.87 & 62.86 & 56.38 & 39.91 & 70.44 & 124.44  \\
RSICCFormer-C~\cite{liu2022remote} & IEEE TGRS 2022 & 83.09 & 74.32 & 66.66 & 62.41 & 38.70 & 73.60 & 132.62 \\
PSNet~\cite{liu2023progressive} & IGARSS 2023 & 83.86 & 75.13 & 67.89 & 62.11 & 38.80 & 73.60 & 132.62 \\
Chg2Cap~\cite{chang2023changes} & IEEE TIP 2023 & 84.43 & 76.35 & 69.12 & 62.98 & 39.42 & 74.34 & 136.25 \\
SEN~\cite{zhou2024single}& IEEE TGRS 2024 & 85.10 & 77.05 & \underline{70.01} & {64.09} & 39.59 & 74.57 & 136.02 \\
SGD-RSCCN~\cite{sun2025scene} & COLING 2025 & 84.17 & 75.16 & 68.05 & 62.48 & 39.18 & 74.24 & 136.20 \\
Diffusion-RSCC~\cite{yu2025diffusion}& IEEE TGRS 2025 & - & - & - & 60.90 & 37.80 & 71.50 & 125.60 \\
{RingMoGPT~\cite{wang2024ringmogpt}} & IEEE TGRS 2025 & 
{83.16} & 
{74.22} & 
{66.74} & 
{60.67} & 
{40.25} & 
{73.97} & 
{135.32} \\

{ChangeChat~\cite{deng2025changechat}} & ICASSP 2025 &  
{83.14} & 
{-} & 
{-} & 
{-} & 
{38.73} & 
{74.01} & 
{136.56} \\

KGBDCNet~\cite{wang2026kgbdcnet} & ISPRS JPRS 2026 & - & - & - & 63.58 & \underline{40.28} & \textbf{75.86} & \underline{139.30} \\

CVMSI-T~\cite{xian2026cross} & TMM 2026 & {-} & {-} & {-} & \textbf{64.83} & {39.72} & {74.97} & {136.59} \\

{DeltaVLM~\cite{deng2026deltavlm}} & Remote Sensing 2026 &
{\underline{85.78}} & 
{\underline{77.15}} & 
{{69.24}} & 
{62.51} & 
{39.47} & 
{{75.01}} & 
{136.72} \\
\hline
\textbf{MGRL-RSCC} &Ours & \textbf{85.89} & \textbf{77.65} & \textbf{70.63} & \underline{64.78} & \textbf{40.61} & \underline{75.56} & \textbf{139.75} \\
\hline
\end{tabular}
}
\end{table*}

\begin{table*}[t]
\centering
\caption{Comparisons with state-of-the-art RSCC methods on the Dubai-CC dataset.}
\label{tab:comparison DUBAI-CC}
\resizebox{0.95\textwidth}{!}{
\begin{tabular}{l|l|ccccccc}
\hline
Method & Publication & BLEU-1 & BLEU-2 & BLEU-3 & BLEU-4
& METEOR & ROUGE-L & CIDEr \\
\hline
DUDA~\cite{park2019robust}
& ICCV 2019
& 58.82 & 43.59 & 33.63 & 25.39 & 22.05 & 48.34 & 62.78 \\
MCCFormers-S~\cite{qiu2021describing}
& ICCV 2021
& 52.97 & 37.02 & 27.62 & 22.57 & 18.64 & 43.29 & 53.81 \\
MCCFormers-D~\cite{qiu2021describing}
& ICCV 2021
& 64.65 & 50.45 & 39.36 & 29.48 & 25.09 & 51.27 & 63.09 \\
RSICCformer-C~\cite{liu2022remote}
& TGRS 2022
& 67.92 & 53.61 & 41.37 & 31.28 & 25.41 & 51.96 & 66.54 \\
Prompt-CC~\cite{liu2023decoupling}
& TGRS 2023
& 70.03 & 58.41 & 49.44 & 40.32 & 26.48 & 55.82 & 85.44 \\
Chg2Cap~\cite{chang2023changes}
& TIP 2023
& 72.04 & 60.18 & 50.84 & 41.70 & 28.92 & 58.66 & 92.49 \\
SEN~\cite{zhou2024single}
& TGRS 2024
& 70.95 & 57.28 & 45.81 & 36.25 & 26.62 & 55.95 & 91.77 \\
SFT~\cite{sun2024lightweight} & JSTARS 2024 & 67.30 & 55.97 & 47.00 & 37.30 & 26.32 & 56.38 & 91.59 \\
Diffusion-RSCC~\cite{yu2025diffusion}
& TGRS 2025
& -- & -- & -- & 33.30 & 27.40 & 56.50 & 88.70 \\
Change3D~\cite{zhu2025change3d} & CVPR 2025 & 72.25 & 58.68 & 47.13 & 36.80 & 27.06 & 56.04 & 86.19 \\
DFM~\cite{wang2026dfm} & arXiv 2026 & 67.86 & 51.01 & 40.61 & 32.08 & 25.75 & 54.36 & 84.65 \\
SAGE-CC~\cite{wang2026sam} & TGRS 2026 & 74.25 & 62.12 & 51.77 & 42.21 & 29.05 & 59.58 & 93.26 \\
\hline
\textbf{MGRL-RSCC}
& {Ours}
& \textbf{76.42}
& \textbf{63.21}
& \textbf{53.09}
& \textbf{43.92}
& \textbf{30.68}
& \textbf{61.03}
& \textbf{99.44} \\
\hline
\end{tabular}
}
\end{table*}

\section{Experiments} \label{sec::experiments}

\subsection{Datasets and Evaluation Metrics}

\noindent $\bullet$ \textbf{LEVIR-CC Dataset.~} 
The LEVIR-CC dataset~\cite{liu2022remote} contains 10,077 pairs of bi-temporal remote sensing images, including 5,038 changed pairs and 5,039 unchanged pairs, which are derived from the LEVIR-CD dataset~\cite{chen2020spatial}. Each image has a spatial resolution of 0.5 m/pixel and a size of 256 $\times$ 256 pixels. The images were acquired from 20 regions in Texas through the Google Earth API, with a temporal gap of 5--14 years between the two acquisitions. Each image pair is annotated with five descriptive sentences, yielding 50,385 captions in total. For unchanged pairs, the annotations are fixed, whereas changed pairs are described with diverse sentences. Following the default experimental setting~\cite{liu2022remote}, we split the dataset into 6,815 pairs for training, 1,333 for validation, and 1,929 for testing.

\noindent $\bullet$ \textbf{Dubai-CC Dataset.~} 
The Dubai-CC dataset~\cite{hoxha2022change} consists of 500 pairs of bi-temporal remote sensing images that capture urbanization changes in Dubai. The images were acquired by the Enhanced Thematic Mapper Plus (ETM+) sensor on board Landsat 7 on May 19, 2000 and June 16, 2010. The original images were cropped into 50 $\times$ 50 pixel tiles, and each pair was annotated with five change descriptions based on Google Maps and publicly available documents, resulting in 2,500 captions in total. Following the default experimental setting~\cite{hoxha2022change}, we split the dataset into 300 pairs for training, 50 for validation, and 150 for testing.

\noindent $\bullet$ \textbf{WHU-CDC Dataset.~}
The WHU-CDC dataset~\cite{shi2024multi} contains 7,434 high-resolution bi-temporal image pairs collected from 2011 to 2016, covering changes in buildings, parking lots, roads, and other categories. In total, the dataset provides 37,170 descriptive sentences. Following the default experimental setting~\cite{shi2024multi}, we split the dataset into 5,947 pairs for training, 743 for validation, and 744 for testing.

\noindent $\bullet$ \textbf{Evaluation Metrics.~}
We adopt four widely used captioning evaluation metrics, including BLEU-N~\cite{papineni2002bleu}, METEOR~\cite{banerjee2005meteor}, ROUGE-L~\cite{lin2004rouge}, and CIDEr~\cite{vedantam2015cider}. BLEU-N evaluates $n$-gram precision between generated and reference captions, reflecting local lexical correspondence at different phrase lengths. METEOR considers both precision and recall together with stemming and synonym matching, providing a complementary measure of lexical and semantic correspondence. ROUGE-L measures the longest common subsequence between generated and reference captions and therefore reflects sequence-level content coverage. CIDEr-D employs TF-IDF-weighted $n$-grams to evaluate the consensus between generated captions and multiple human references, while emphasizing informative expressions.


\subsection{Implementation Details}
The proposed deep learning methods are implemented in PyTorch~\cite{paszke2019pytorch}. All training and evaluation experiments are carried out on a single NVIDIA RTX 4090 GPU with 24 GB of memory. For optimization, we use Adam for all trainable modules. The learning rate is set to $1\times10^{-4}$ for the encoder and attentive encoder, and $1\times10^{-5}$ for the decoder. The learning rate is decayed by a factor of 0.5 every 5 epochs. In the unified reward setting, the decoder learning rate is additionally boosted once to $2\times10^{-5}$ at epoch 30. The model is trained for at most 50 epochs. After each training epoch, validation is performed on the development set, and the checkpoint selected according to the validation BLEU-4 score is used for final testing.
More details can be found in our source code.

\subsection{Comparison on Public Benchmark Datasets}

As shown in Tables~\ref{tab:comparison LEVIR-CC}, \ref{tab:comparison DUBAI-CC}, and \ref{tab:comparison WHU-CDC}, we compare \textsc{MGRL-RSCC} with representative remote sensing change captioning (RSCC) methods on three public benchmark datasets, namely LEVIR-CC, Dubai-CC, and WHU-CDC. The compared methods are evaluated using BLEU-1, BLEU-2, BLEU-3, BLEU-4, METEOR, ROUGE-L, and CIDEr, with higher values indicating better performance.

On the LEVIR-CC dataset, \textsc{MGRL-RSCC} achieves the best results on five of the seven evaluation metrics and the second-best results on BLEU-4 and ROUGE-L. Specifically, it obtains BLEU-1, BLEU-2, BLEU-3, and BLEU-4 scores of 85.89, 77.65, 70.63, and 64.78, respectively, together with a METEOR score of 40.61, a ROUGE-L score of 75.56, and a CIDEr score of 139.75. Compared with the strongest competing result for each metric, \textsc{MGRL-RSCC} improves BLEU-1, BLEU-2, BLEU-3, METEOR, and CIDEr by 0.11, 0.50, 0.62, 0.33, and 0.45 points, respectively. Meanwhile, its BLEU-4 and ROUGE-L scores are only 0.05 and 0.30 points lower than the corresponding best results. These results demonstrate that \textsc{MGRL-RSCC} achieves strong and balanced performance across different captioning evaluation metrics.

On the Dubai-CC dataset, \textsc{MGRL-RSCC} consistently outperforms all compared methods across all seven metrics. It achieves BLEU-1, BLEU-2, BLEU-3, and BLEU-4 scores of 76.42, 63.21, 53.09, and 43.92, respectively, as well as a METEOR score of 30.68, a ROUGE-L score of 61.03, and a CIDEr score of 99.44. Compared with the strongest competing result for each metric, \textsc{MGRL-RSCC} improves BLEU-1, BLEU-2, BLEU-3, BLEU-4, METEOR, ROUGE-L, and CIDEr by 2.17, 1.09, 1.32, 1.71, 1.63, 1.45, and 6.18 points, respectively. Notably, the 6.18-point improvement in CIDEr indicates that the generated captions achieve stronger consensus with the reference descriptions. The consistent improvements across all evaluation metrics further demonstrate the strong performance of \textsc{MGRL-RSCC} under the relatively small-scale training setting of Dubai-CC.

On the WHU-CDC dataset, \textsc{MGRL-RSCC} achieves the best performance on all seven evaluation metrics. Specifically, it obtains BLEU-1, BLEU-2, BLEU-3, and BLEU-4 scores of 86.33, 81.46, 77.81, and 75.00, respectively. It also achieves a METEOR score of 48.13, a ROUGE-L score of 81.66, and a CIDEr score of 156.88. Compared with the strongest competing result for each metric, \textsc{MGRL-RSCC} improves BLEU-1, BLEU-2, BLEU-3, BLEU-4, METEOR, ROUGE-L, and CIDEr by 0.29, 0.30, 0.53, 0.58, 0.24, 0.19, and 0.67 points, respectively. Although the improvements on some metrics are relatively modest, the consistent gains across all seven evaluation metrics demonstrate the strong and stable performance of \textsc{MGRL-RSCC} on the WHU-CDC dataset.

Overall, \textsc{MGRL-RSCC} achieves consistently competitive performance across the three benchmark datasets with different data scales and scene characteristics, demonstrating the general effectiveness of the proposed framework for remote sensing change captioning. The contributions of the different reward components are further investigated in the following ablation studies.

\begin{table*}[t]
\centering
\caption{Comparisons with state-of-the-art RSCC methods on the WHU-CDC dataset.}
\label{tab:comparison WHU-CDC}
\resizebox{0.95\textwidth}{!}{
\begin{tabular}{l|l|ccccccc}
\hline
Method & Publication & BLEU-1 & BLEU-2 & BLEU-3 & BLEU-4
& METEOR & ROUGE-L & CIDEr \\
\hline
DUDA~\cite{park2019robust}
& ICCV 2019
& 79.04 & 69.53 & 61.57 & 55.64 & 34.29 & 68.98 & 121.85 \\

MCCFormers-S~\cite{qiu2021describing}
& ICCV 2021
& 82.14 & 76.29 & 71.08 & 66.51 & 43.50 & 79.76 & 148.88 \\

MCCFormers-D~\cite{qiu2021describing}
& ICCV 2021
& 73.29 & 67.88 & 64.03 & 60.96 & 39.69 & 73.67 & 134.92 \\

RSICCformer-C~\cite{liu2022remote}
& TGRS 2022
& 78.25 & 72.82 & 68.57 & 65.14 & 44.35 & 76.50 & 143.44 \\

MaskApproxNet~\cite{sun2025mask}
& TGRS 2025
& 81.34 & 75.68 & 71.16 & 67.73 & 43.89 & 75.41 & 135.31 \\

CTMTNet~\cite{shi2024multi}
& TGRS 2024
& 83.56 & 77.66 & 72.76 & 69.00 & 45.39 & 79.23 & 149.40 \\

Semantic-CC~\cite{zhu2024semantic}
& TGRS 2024
& 82.77 & 76.32 & 71.59 & 68.43 & 44.49 & 78.23 & 150.23 \\

KCFI~\cite{yang2025enhancing}
& TIP 2025
& 83.34 & 77.27 & 72.40 & 68.47 & 44.95 & 79.59 & 149.32 \\

CTM~\cite{bai2025cross}
& JSTARS 2025
& 85.36 & 79.49 & 75.36 & 72.36 & 46.98 & 80.97 & 153.29 \\

PTNet~\cite{gao2026uav}
& arXiv 2026
& 83.94 & 77.89 & 72.94 & 69.37 & 45.69 & 79.64 & 150.02 \\

D3-Net~\cite{peng2026frequency} & JSTARS 2026 & 85.42 & 80.02 & 76.23 & 73.15 & 47.13 & 81.47 & 154.33 \\

SAGE-CC~\cite{wang2026sam}
& TGRS 2026
& 86.04 & 81.16 & 77.28 & 74.42 & 47.89 & 80.90 & 156.21 \\
\hline

\textbf{MGRL-RSCC}
& Ours
& \textbf{86.33}
& \textbf{81.46}
& \textbf{77.81}
& \textbf{75.00}
& \textbf{48.13}
& \textbf{81.66}
& \textbf{156.88} \\
\hline
\end{tabular}
}
\end{table*}

\begin{table*}
\centering
\caption{Progressive ablation study on the LEVIR-CC dataset. `MR`, `CR`, and `SR` denote the metric rewards, the change-state reward, and the structural-semantic rewards, respectively. `Baseline` denotes CE-only training. Symbols ``$\times$'' and ``$\checkmark$'' indicate whether a reward group is excluded or included. Higher values indicate better performance, and the best results are highlighted in bold.}
\label{tab:ablation_progressive}
\resizebox{0.95\textwidth}{!}{
\begin{tabular}{c|ccc|ccccccc}
\hline
Method & MR & CR & SR & BLEU-1 & BLEU-2 & BLEU-3 & BLEU-4 & METEOR & ROUGE-L & CIDEr \\
\hline
Baseline & $\times$ & $\times$ & $\times$ & 84.43 & 76.35 & 69.12 & 62.98 & 39.42 & 74.34 & 136.25 \\
(a) & $\checkmark$ & $\times$ & $\times$ & 84.71 & 76.57 & 69.61 & 63.54 & 39.85 & 74.58 & 138.08 \\
(b) & $\checkmark$ & $\checkmark$ & $\times$ & 84.87 & 76.73 & 69.78 & 63.71 & 39.98 & 74.72 & 137.91 \\
(c) & $\checkmark$ & $\times$ & $\checkmark$ & 85.42 & 77.05 & 70.11 & 64.56 & 40.23 & 74.95 & 138.64 \\
\hline
(d) & $\checkmark$ & $\checkmark$ & $\checkmark$ & \textbf{85.89} & \textbf{77.65} & \textbf{70.63} & \textbf{64.78} & \textbf{40.61} & \textbf{75.56} & \textbf{139.75} \\
\hline
\end{tabular}
}
\end{table*}




\begin{table*}
\centering
\caption{Ablation study on the top-$k$ sampling size during reinforcement learning on the LEVIR-CC dataset.}
\resizebox{0.95\textwidth}{!}{
\tiny
\begin{tabular}{c|ccccccc}
\hline
Top-$k$ & BLEU-1 & BLEU-2 & BLEU-3 & BLEU-4 & METEOR & ROUGE-L & CIDEr \\
\hline
2 & 84.74 & 76.21 & 69.22 & 63.39 & 40.29 & 74.76 & 139.04 \\
3 & \textbf{85.89} & \textbf{77.65} & \textbf{70.63} & \textbf{64.78} & \textbf{40.61} & \textbf{75.56} & \textbf{139.75} \\
4 & 84.21 & 75.96 & 68.94 & 63.02 & 39.19 & 73.81 & 135.86 \\
\hline
\end{tabular}
}
\label{tab:topk_comparison}
\end{table*}

\begin{table*}
\centering
\caption{Ablation study on the sampling temperature during reinforcement learning on the LEVIR-CC dataset.}
\label{tab:temperature_comparison}
\resizebox{0.95\textwidth}{!}{
\tiny
\begin{tabular}{c|ccccccc}
\hline
Temperature & BLEU-1 & BLEU-2 & BLEU-3 & BLEU-4 & METEOR & ROUGE-L & CIDEr \\
\hline
0.6 & 85.56 & 77.37 & 70.62 & \textbf{65.03} & 40.33 & 75.18 & 138.65 \\
0.8 & \textbf{85.89} & \textbf{77.65} & \textbf{70.63} & 64.78 & \textbf{40.61} & \textbf{75.56} & \textbf{139.75} \\
1.0 & 85.20 & 77.04 & 70.28 & 64.66 & 39.96 & 74.77 & 137.50 \\
\hline
\end{tabular}
}
\end{table*}

\subsection{Ablation Study}

\noindent $\bullet$ \textbf{Component Analysis.~} As shown in Table~\ref{tab:ablation_progressive}, we systematically evaluate the contributions of the three reward groups in MGRL-RSCC, namely the metric rewards (MR), the change-state reward (CR), and the structural-semantic rewards (SR). Starting from the CE-only baseline, introducing MR in Method (a) consistently improves all evaluation metrics, increasing BLEU-1 from 84.43 to 84.71, BLEU-2 from 76.35 to 76.57, BLEU-3 from 69.12 to 69.61, BLEU-4 from 62.98 to 63.54, METEOR from 39.42 to 39.85, ROUGE-L from 74.34 to 74.58, and CIDEr from 136.25 to 138.08. These improvements indicate that directly incorporating sequence-level metric feedback helps reduce the mismatch between token-level supervised optimization and sequence-level caption evaluation. When CR is further introduced in Method (b), six of the seven metrics continue to improve over Method (a), while CIDEr decreases slightly from 138.08 to 137.91. This result suggests that the change-state reward provides complementary supervision for global change-state correctness, although its effect is not uniformly reflected by all conventional captioning metrics. Compared with Method (a), incorporating SR in Method (c) yields more pronounced improvements than incorporating CR in Method (b), reaching 85.42, 77.05, 70.11, and 64.56 on BLEU-1 through BLEU-4, respectively, together with 40.23 METEOR, 74.95 ROUGE-L, and 138.64 CIDEr. This result suggests that structural-semantic supervision provides substantial complementary benefits by encouraging the model to better preserve fine-grained object, relational, and contextual information. Finally, Method (d), which jointly incorporates MR, CR, and SR, achieves the best performance across all seven evaluation metrics, reaching 85.89 BLEU-1, 77.65 BLEU-2, 70.63 BLEU-3, 64.78 BLEU-4, 40.61 METEOR, 75.56 ROUGE-L, and 139.75 CIDEr.

Overall, the ablation results demonstrate that the three reward groups provide complementary optimization signals. MR aligns reinforcement learning with sequence-level caption evaluation, CR introduces explicit supervision for global change-state correctness, and SR further constrains fine-grained structural-semantic information. Their joint use yields the strongest overall performance. Compared with the CE-only baseline, the full model improves BLEU-4 by 1.80 points, METEOR by 1.19 points, and CIDEr by 3.50 points, while also achieving consistent gains on the remaining metrics. These results validate the effectiveness and complementarity of the proposed multi-granularity reward design.

\begin{figure}
\centering
\includegraphics[width=\linewidth]{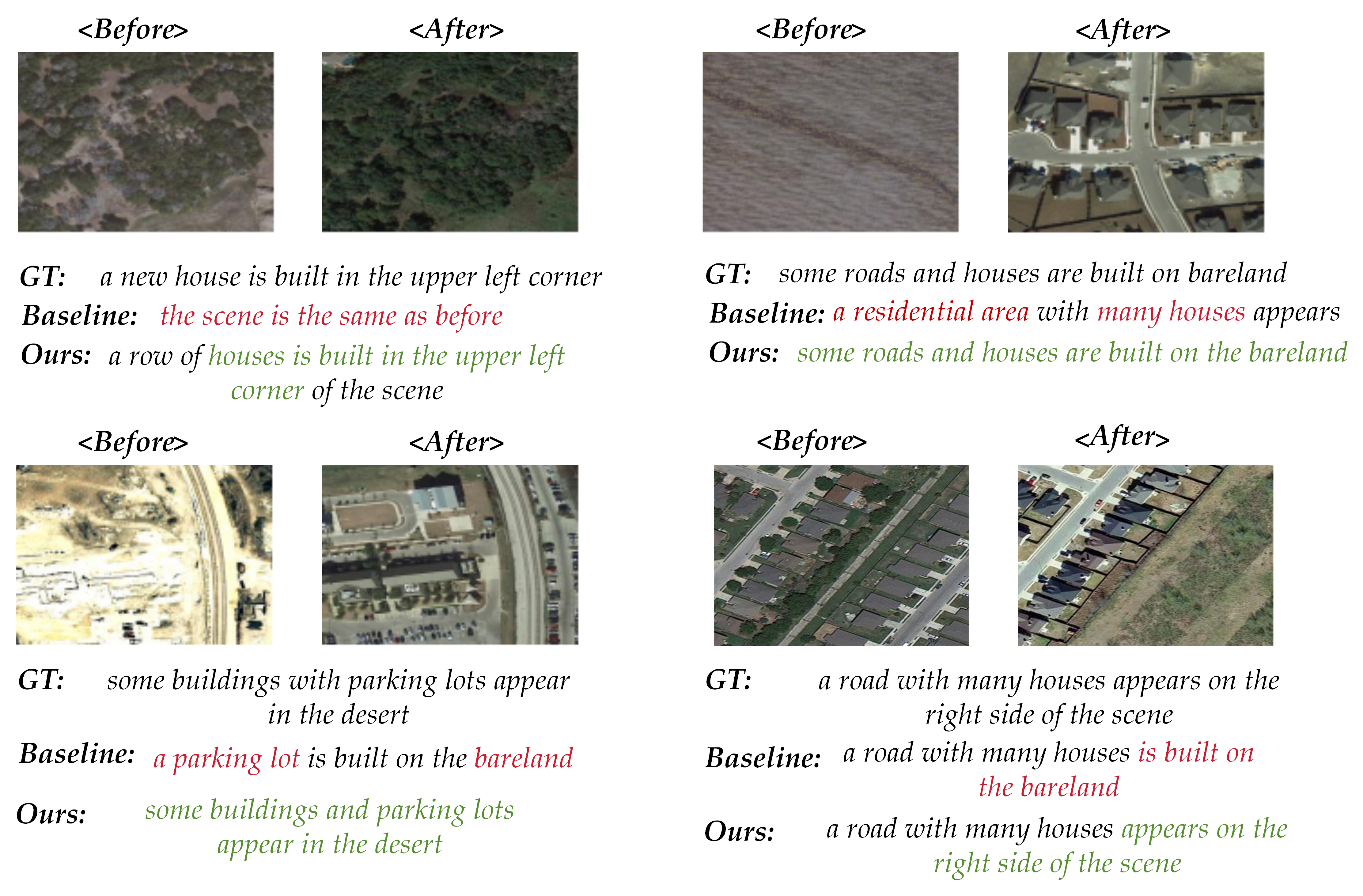}
\caption{Qualitative comparisons between the baseline and MGRL-RSCC on the LEVIR-CC dataset. Correct and incorrect expressions are highlighted in green and red, respectively.} 
\label{fig:captioning results} 
\end{figure}

\begin{figure*}
\centering
\includegraphics[width=1\textwidth]{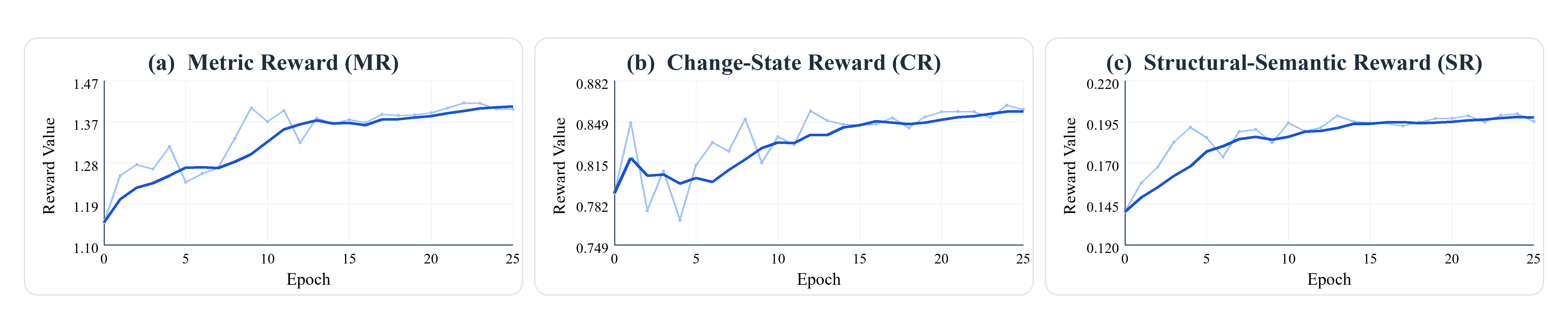}
\caption{Reward values on the validation set during reinforcement learning. Subfigures (a)--(c) show the evolution of the metric reward (MR), change-state reward (CR), and structural-semantic reward (SR), respectively. All three reward signals exhibit generally stable and convergent trends throughout training.}
\label{fig:reward_pic}
\end{figure*}

\noindent $\bullet$ \textbf{Analysis of the Top-$k$ Sampling Size.~} 
As shown in Table~\ref{tab:topk_comparison}, we investigate the effect of the top-$k$ sampling size on the performance of MGRL-RSCC during reinforcement learning. Among the three settings, $k=3$ achieves the best performance across all evaluation metrics, indicating that a moderately sized candidate set provides a favorable balance between sampling diversity and reliability. When $k=2$, the restricted candidate set may limit the exploration of alternative caption sequences, resulting in relatively lower performance. In contrast, increasing the sampling size to $k=4$ leads to a consistent performance degradation. A possible reason is that a larger candidate set includes more relatively low-probability tokens during sampling, thereby increasing the uncertainty of the sampled caption sequences. Based on these results, we adopt $k=3$ in all experiments.

\noindent $\bullet$ \textbf{Analysis of the Sampling Temperature.~} 
As shown in Table~\ref{tab:temperature_comparison}, we further investigate the effect of the sampling temperature on the performance of MGRL-RSCC during reinforcement learning. A lower temperature produces a sharper probability distribution and increases the relative probability of high-confidence tokens during sampling. Accordingly, a temperature of 0.6 achieves the highest BLEU-4 score, suggesting that a relatively concentrated sampling distribution may favor local $n$-gram matching. In comparison, a temperature of 0.8 achieves the best performance on six of the seven evaluation metrics, including BLEU-1, BLEU-2, BLEU-3, METEOR, ROUGE-L, and CIDEr, indicating a better overall balance between sampling diversity and generation reliability. When the temperature is further increased to 1.0, the performance decreases across all seven metrics. This may be because a flatter probability distribution increases the probability of selecting relatively low-confidence candidates during sampling, thereby introducing greater uncertainty into reward-based optimization. Considering the overall performance across different evaluation metrics, we adopt a sampling temperature of 0.8 in all experiments.

\begin{figure}[t]
\centering
\includegraphics[width=\linewidth]{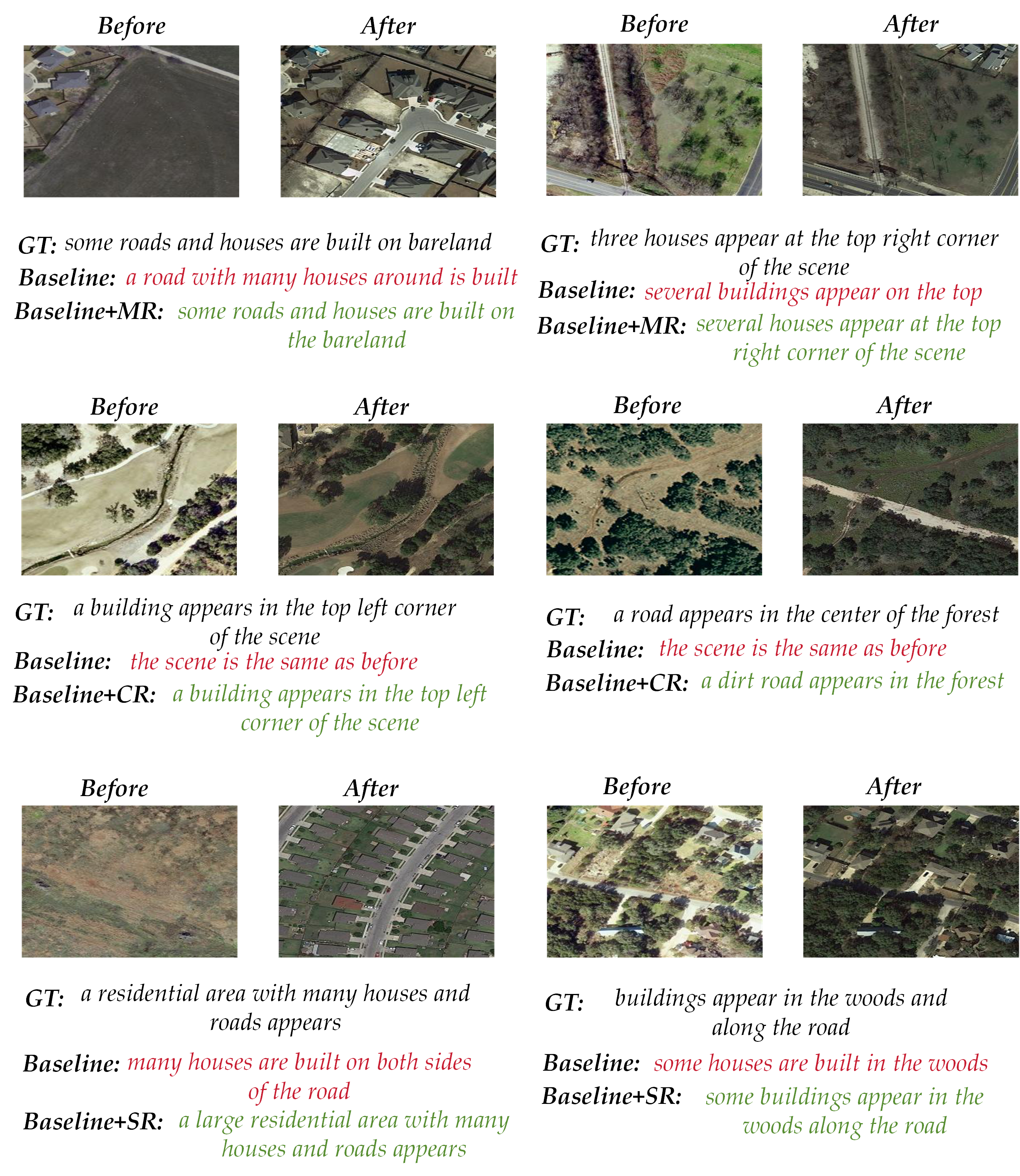}
\caption{Qualitative comparison of captions generated with different reward signals. MR improves linguistic alignment, CR enhances change-state correctness, and SR strengthens fine-grained structural-semantic consistency.}
\label{fig:reward_signal_results}
\end{figure}

\subsection{Efficiency Analysis}

As shown in Table~\ref{tab:MGRL-RSCC_efficiency}, we compare the parameter scale and captioning performance of MGRL-RSCC with representative RSCC methods on the LEVIR-CC dataset. MGRL-RSCC contains 329.03M parameters, which is larger than MCCFormers-S and RSICCFormer-C but smaller than PSNet, Prompt-CC, Sparse Focus, and RSCaMa. In our implementation, MGRL-RSCC requires 107.3 GFLOPs for greedy inference and has an inference latency of 137.1 ms. Despite its moderate parameter scale, MGRL-RSCC achieves a BLEU-4 score of 64.78 and the highest CIDEr score of 139.75 among the compared methods.

It is worth noting that the proposed multi-granularity reward learning is applied only during training and does not introduce additional network modules or computation into the inference architecture. Therefore, the performance improvement is achieved without increasing the inference-time model complexity relative to the corresponding captioning architecture. These results indicate that the gains of MGRL-RSCC are primarily attributed to the proposed reward-based optimization rather than simply increasing the model size. Overall, MGRL-RSCC achieves a favorable balance between model scale and captioning performance.

\subsection{Visualization} 

\noindent $\bullet$ \textbf{Captioning Results.~} As shown in Figure~\ref{fig:captioning results}, we provide several qualitative examples to illustrate the effectiveness of our proposed MGRL-RSCC model for remote sensing change captioning. For each bi-temporal remote sensing image pair, we compare the captions generated by MGRL-RSCC with those produced by the baseline model, together with the ground-truth annotations. To make the comparison more intuitive, the inaccurate words or phrases in the baseline captions are highlighted in red, while the words or phrases in the captions generated by MGRL-RSCC that are consistent with the ground truth are highlighted in green. From these examples, it can be observed that MGRL-RSCC generally produces captions that are more accurate and more consistent with the ground-truth semantic changes than the baseline model.

\noindent $\bullet$ \textbf{Effects of Different Reward Signals.~}
As shown in Figure~\ref{fig:reward_signal_results}, we present several qualitative examples to illustrate the complementary effects of the proposed metric reward (MR), change-state reward (CR), and structural-semantic reward (SR). For each bi-temporal image pair, the caption generated by the baseline is compared with that produced after introducing the corresponding reward signal, together with the ground-truth annotation. MR encourages closer linguistic alignment with the reference captions, leading to more accurate object, quantity, and location expressions. CR improves the recognition of the global change state and reduces false no-change predictions when actual changes occur. SR further promotes the generation of fine-grained details and coherent object-relation-context structures. These examples demonstrate that the three reward signals provide targeted supervision at complementary semantic levels and jointly support more accurate and semantically complete change descriptions.


\begin{table}[H]
\centering
\caption{Efficiency and performance comparison on the LEVIR-CC dataset.}
\small
\label{tab:MGRL-RSCC_efficiency}
\begin{tabular}{l|ccc}
\hline
\textbf{Method} & \textbf{Params(M)} & \textbf{BLEU-4} & \textbf{CIDEr} \\
\hline
MCCFormers-S   & 69.86  & 59.07 & 122.06 \\
RSICCFormer-C  & 172.80 & 62.41 & 132.62 \\
PSNet          & 424.39 & 62.11 & 132.62 \\
RSCaMa         & 790.18 & 65.24 & 136.56 \\
Prompt-CC      & 408.58 & 63.54 & 136.44 \\
Sparse Focus   & 687.10 & 62.87 & 137.05 \\
\hline
MGRL-RSCC  & 329.03 & 64.78 & 139.75 \\
\hline
\end{tabular}
\end{table}

\noindent $\bullet$ \textbf{Reward Dynamics.~}
As shown in Figure~\ref{fig:reward_pic}, the validation reward curves remain generally stable and exhibit clear convergence trends during reinforcement learning. The metric reward in Figure~\ref{fig:reward_pic}(a) increases rapidly in the early stage and then gradually stabilizes, indicating that reinforcement learning progressively improves the linguistic quality of the generated captions. The change-state reward in Figure~\ref{fig:reward_pic}(b) also shows an overall upward trend, suggesting that the model becomes more reliable in distinguishing changed scenes from unchanged ones. Meanwhile, the structural-semantic reward in Figure~\ref{fig:reward_pic}(c), which jointly considers fine-grained details and structured semantic consistency, steadily improves and eventually converges. Overall, these curves demonstrate that the proposed reward components provide stable optimization signals and facilitate a smooth reinforcement learning process.

\begin{figure}[t]
\centering
\includegraphics[width=\linewidth]{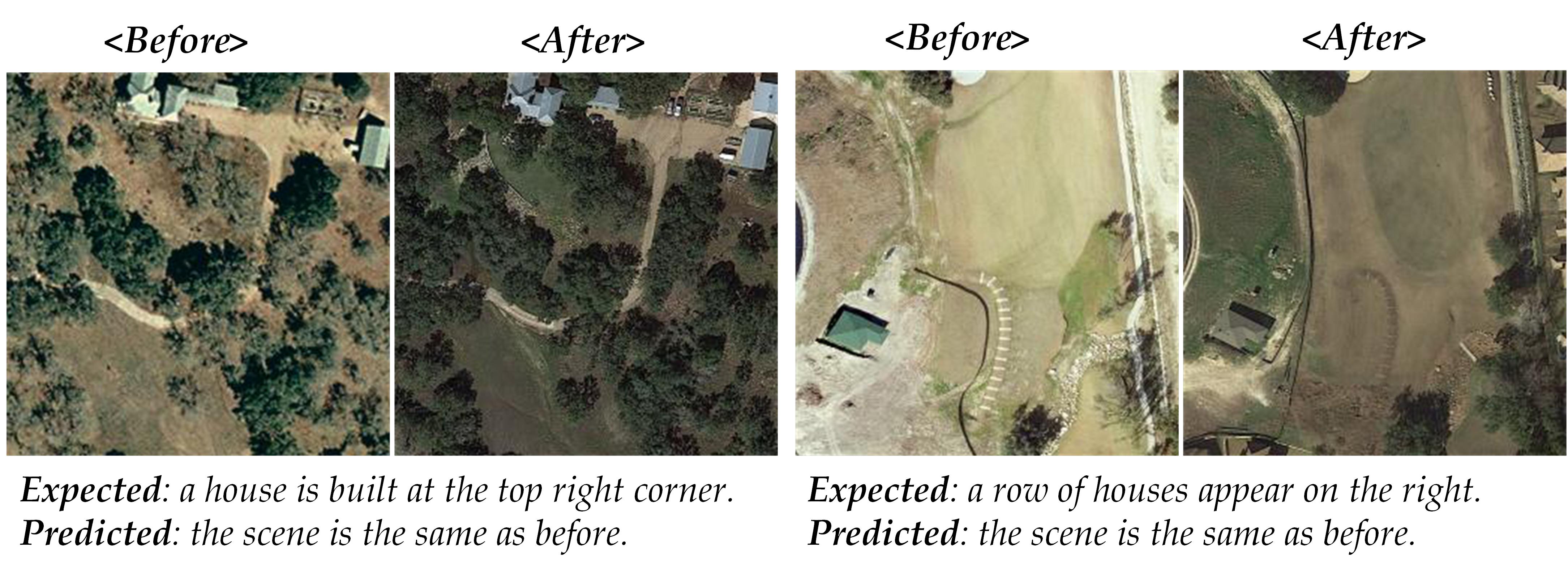}
\caption{Visualization of representative failure cases.}
\label{fig:failure_vis}
\end{figure}

\subsection{Limitation Analysis}
Although MGRL-RSCC improves caption generation by jointly optimizing caption-level metric rewards, change-state consistency, and structural-semantic constraints, several limitations remain. As shown in Figure~\ref{fig:failure_vis}, MGRL-RSCC may still overlook visually subtle or spatially marginal changes. In the first example, a newly built house located in the upper-right corner is relatively small and surrounded by dense vegetation. Moreover, the two temporal images exhibit noticeable appearance differences, making the changed object difficult to distinguish from irrelevant environmental variations. In the second example, a newly constructed row of villas appears only along the right boundary of the image and is partially visible. In both cases, MGRL-RSCC incorrectly predicts that the scene remains unchanged.

These failure cases indicate that the change-state reward can reduce, but cannot completely eliminate, incorrect no-change predictions. In particular, when subtle objects or boundary regions are insufficiently represented by the visual encoder, the structural-semantic rewards cannot fully compensate for the missing visual evidence. Moreover, the detail and knowledge-graph rewards rely on predefined phrase normalization rules, semantic tag sets, and a finite collection of graph triplets. Rare object categories, unseen relations, and diverse linguistic expressions may therefore receive incomplete reward supervision, limiting the generalization ability of MGRL-RSCC across different datasets and change patterns. Future work will explore spatially grounded and open-vocabulary reward modeling to better capture subtle changes and unseen semantic relations.

\section{Conclusion} \label{sec::conclusion} 
In this paper, we propose MGRL-RSCC, a unified reinforcement learning framework for remote sensing change captioning. The proposed method combines bi-temporal visual encoding and Transformer-based caption generation with a multi-level reward function consisting of caption-level metric rewards, a scene-level change-state reward, and fine-grained structural-semantic rewards. By jointly optimizing linguistic quality, global change correctness, detailed change information, and knowledge-graph consistency through self-critical sequence training, MGRL-RSCC generates more accurate and semantically consistent change descriptions. Extensive experiments and ablation studies on public benchmark datasets demonstrate the competitive performance of MGRL-RSCC and verify the effectiveness of the proposed reward components.

\section*{Acknowledgment} \label{sec::Acknowledgment} 
This work was supported by the National Natural Science Foundation of China under Grant 62572004, 62102205, U24A20342. Anhui Provincial Natural Science Foundation-Outstanding Youth Project, 2408085Y032. The authors acknowledge the High-performance Computing Platform of Anhui University for providing computing resources.


\small{ 
\bibliographystyle{IEEEtran}
\bibliography{reference}

\begin{thebibliography}{10}
\providecommand{\url}[1]{#1}
\csname url@samestyle\endcsname
\providecommand{\newblock}{\relax}
\providecommand{\bibinfo}[2]{#2}
\providecommand{\BIBentrySTDinterwordspacing}{\spaceskip=0pt\relax}
\providecommand{\BIBentryALTinterwordstretchfactor}{4}
\providecommand{\BIBentryALTinterwordspacing}{\spaceskip=\fontdimen2\font plus
\BIBentryALTinterwordstretchfactor\fontdimen3\font minus
  \fontdimen4\font\relax}
\providecommand{\BIBforeignlanguage}[2]{{%
\expandafter\ifx\csname l@#1\endcsname\relax
\typeout{** WARNING: IEEEtran.bst: No hyphenation pattern has been}%
\typeout{** loaded for the language `#1'. Using the pattern for}%
\typeout{** the default language instead.}%
\else
\language=\csname l@#1\endcsname
\fi
#2}}
\providecommand{\BIBdecl}{\relax}
\BIBdecl

\bibitem{hoxha2022change}
G.~Hoxha, S.~Chouaf, F.~Melgani, and Y.~Smara, ``Change captioning: A new
  paradigm for multitemporal remote sensing image analysis,'' \emph{IEEE
  Transactions on Geoscience and Remote Sensing}, vol.~60, pp. 1--14, 2022.

\bibitem{liu2022remote}
C.~Liu, R.~Zhao, H.~Chen, Z.~Zou, and Z.~Shi, ``Remote sensing image change
  captioning with dual-branch transformers: A new method and a large scale
  dataset,'' \emph{IEEE Transactions on Geoscience and Remote Sensing},
  vol.~60, pp. 1--20, 2022.

\bibitem{zou2025remote}
S.~Zou, Y.~Wei, Y.~Xie, M.~Lao, and X.~Luan, ``Remote sensing image change
  captioning: A comprehensive review: S. zou et al.'' \emph{International
  Journal of Multimedia Information Retrieval}, vol.~14, no.~3, p.~26, 2025.

\bibitem{chang2023changes}
S.~Chang and P.~Ghamisi, ``Changes to captions: An attentive network for remote
  sensing change captioning,'' \emph{IEEE Transactions on Image Processing},
  vol.~32, pp. 6047--6060, 2023.

\bibitem{liu2023progressive}
C.~Liu, J.~Yang, Z.~Qi, Z.~Zou, and Z.~Shi, ``Progressive scale-aware network
  for remote sensing image change captioning,'' in \emph{IGARSS 2023-2023 IEEE
  International Geoscience and Remote Sensing Symposium}.\hskip 1em plus 0.5em
  minus 0.4em\relax IEEE, 2023, pp. 6668--6671.

\bibitem{zhu2024semantic}
Y.~Zhu, L.~Li, K.~Chen, C.~Liu, F.~Zhou, and Z.~Shi, ``Semantic-cc: Boosting
  remote sensing image change captioning via foundational knowledge and
  semantic guidance,'' \emph{IEEE Transactions on Geoscience and Remote
  Sensing}, vol.~62, pp. 1--16, 2024.

\bibitem{yang2025enhancing}
C.~Yang, Z.~Li, H.~Jiao, Z.~Gao, and L.~Zhang, ``Enhancing perception of key
  changes in remote sensing image change captioning,'' \emph{IEEE Transactions
  on Image Processing}, 2025.

\bibitem{deng2025changechat}
P.~Deng, W.~Zhou, and H.~Wu, ``Changechat: An interactive model for remote
  sensing change analysis via multimodal instruction tuning,'' in \emph{ICASSP
  2025-2025 IEEE International Conference on Acoustics, Speech and Signal
  Processing (ICASSP)}.\hskip 1em plus 0.5em minus 0.4em\relax IEEE, 2025, pp.
  1--5.

\bibitem{zhang2026rsc}
Y.~Zhang, B.~Wang, Z.~Shao, W.~Chen, and W.~Zhao, ``Rsc-cot: Visual-cot
  reasoning and reinforced optimization for remote sensing change captioning,''
  in \emph{ICASSP 2026-2026 IEEE International Conference on Acoustics, Speech
  and Signal Processing (ICASSP)}.\hskip 1em plus 0.5em minus 0.4em\relax IEEE,
  2026, pp. 11\,272--11\,276.

\bibitem{liu2024rscama}
C.~Liu, K.~Chen, B.~Chen, H.~Zhang, Z.~Zou, and Z.~Shi, ``Rscama: Remote
  sensing image change captioning with state space model,'' \emph{IEEE
  Geoscience and Remote Sensing Letters}, vol.~21, pp. 1--5, 2024.

\bibitem{karaca2025robust}
A.~C. Karaca, E.~Ozelbas, S.~Berber, O.~Karimli, T.~Yildirim, and M.~F.
  Amasyali, ``Robust change captioning in remote sensing: Second-cc dataset and
  mmodalcc framework,'' \emph{IEEE Journal of Selected Topics in Applied Earth
  Observations and Remote Sensing}, 2025.

\bibitem{xue2026towards}
J.~Xue, Q.~Deng, X.~Wu, K.~Yao, X.~Yin, F.~Yu, W.~Zhou, Y.~Zhong, Y.~Liu, and
  D.~Yang, ``Towards comprehensive interactive change understanding in remote
  sensing: A large-scale dataset and dual-granularity enhanced vlm,''
  \emph{IEEE Transactions on Geoscience and Remote Sensing}, 2026.

\bibitem{bengio2015scheduled}
S.~Bengio, O.~Vinyals, N.~Jaitly, and N.~Shazeer, ``Scheduled sampling for
  sequence prediction with recurrent neural networks,'' \emph{Advances in
  neural information processing systems}, vol.~28, 2015.

\bibitem{rennie2017self}
S.~J. Rennie, E.~Marcheret, Y.~Mroueh, J.~Ross, and V.~Goel, ``Self-critical
  sequence training for image captioning,'' in \emph{Proceedings of the IEEE
  conference on computer vision and pattern recognition}, 2017, pp. 7008--7024.

\bibitem{wei2022chain}
J.~Wei, X.~Wang, D.~Schuurmans, M.~Bosma, F.~Xia, E.~Chi, Q.~V. Le, D.~Zhou
  \emph{et~al.}, ``Chain-of-thought prompting elicits reasoning in large
  language models,'' \emph{Advances in neural information processing systems},
  vol.~35, pp. 24\,824--24\,837, 2022.

\bibitem{shi2024multi}
J.~Shi, M.~Zhang, Y.~Hou, R.~Zhi, and J.~Liu, ``A multi-task network and two
  large scale datasets for change detection and captioning in remote sensing
  images,'' \emph{IEEE Transactions on Geoscience and Remote Sensing}, 2024.

\bibitem{cai2023interactive}
C.~Cai, Y.~Wang, and K.-H. Yap, ``Interactive change-aware transformer network
  for remote sensing image change captioning,'' \emph{Remote Sensing}, vol.~15,
  no.~23, p. 5611, 2023.

\bibitem{sun2024lightweight}
D.~Sun, Y.~Bao, J.~Liu, and X.~Cao, ``A lightweight sparse focus transformer
  for remote sensing image change captioning,'' \emph{IEEE Journal of Selected
  Topics in Applied Earth Observations and Remote Sensing}, vol.~17, pp.
  18\,727--18\,738, 2024.

\bibitem{chen2026rscc}
Z.~Chen, C.~Wang, N.~Zhang, and F.~Zhang, ``Rscc: A large-scale remote sensing
  change caption dataset for disaster events,'' \emph{Advances in Neural
  Information Processing Systems}, vol.~38, 2026.

\bibitem{qu2026mask}
Y.~Qu and H.~Zhang, ``A mask-guided multigranular mamba network for remote
  sensing change captioning,'' \emph{Remote Sensing}, vol.~18, no.~7, p. 1048,
  2026.

\bibitem{zhao2026disturbance}
Y.~Zhao, S.~Lei, H.-C. Li, T.~Celik, and J.~Pan, ``Disturbance-robust remote
  sensing change captioning with self-supervised multifrequency
  representation,'' \emph{Journal of Remote Sensing}, vol.~6, p. 1037, 2026.

\bibitem{zhang2017actor}
L.~Zhang, F.~Sung, F.~Liu, T.~Xiang, S.~Gong, Y.~Yang, and T.~M. Hospedales,
  ``Actor-critic sequence training for image captioning,'' \emph{arXiv preprint
  arXiv:1706.09601}, 2017.

\bibitem{shen2020remote}
X.~Shen, B.~Liu, Y.~Zhou, J.~Zhao, and M.~Liu, ``Remote sensing image
  captioning via variational autoencoder and reinforcement learning,''
  \emph{Knowledge-Based Systems}, vol. 203, p. 105920, 2020.

\bibitem{chavhan2021novel}
R.~Chavhan, B.~Banerjee, X.~X. Zhu, and S.~Chaudhuri, ``A novel actor
  dual-critic model for remote sensing image captioning,'' in \emph{2020 25th
  International Conference on Pattern Recognition (ICPR)}.\hskip 1em plus 0.5em
  minus 0.4em\relax IEEE, 2021, pp. 4918--4925.

\bibitem{zhang2025sc}
L.~Zhang, X.~Zeng, K.~Li, G.~Yu, and T.~Chen, ``Sc-captioner: Improving image
  captioning with self-correction by reinforcement learning,'' in
  \emph{Proceedings of the IEEE/CVF International Conference on Computer
  Vision}, 2025, pp. 23\,145--23\,155.

\bibitem{xing2025caprl}
L.~Xing, X.~Dong, Y.~Zang, Y.~Cao, J.~Liang, Q.~Huang, J.~Wang, F.~Wu, and
  D.~Lin, ``Caprl: Stimulating dense image caption capabilities via
  reinforcement learning,'' \emph{arXiv preprint arXiv:2509.22647}, 2025.

\bibitem{tang2026cccaption}
Z.~Tang, L.~Wang, J.~Qi, W.~Jiang, P.~Hou, A.~Zeng, and J.~Huang, ``Cccaption:
  Dual-reward reinforcement learning for complete and correct image
  captioning,'' in \emph{Proceedings of the IEEE/CVF Conference on Computer
  Vision and Pattern Recognition}, 2026, pp. 22\,153--22\,163.

\bibitem{yang2019auto}
X.~Yang, K.~Tang, H.~Zhang, and J.~Cai, ``Auto-encoding scene graphs for image
  captioning,'' in \emph{Proceedings of the IEEE/CVF conference on computer
  vision and pattern recognition}, 2019, pp. 10\,685--10\,694.

\bibitem{nguyen2021defense}
K.~Nguyen, S.~Tripathi, B.~Du, T.~Guha, and T.~Q. Nguyen, ``In defense of scene
  graphs for image captioning,'' in \emph{Proceedings of the IEEE/CVF
  international conference on computer vision}, 2021, pp. 1407--1416.

\bibitem{zhang2021image}
Y.~Zhang, X.~Shi, S.~Mi, and X.~Yang, ``Image captioning with transformer and
  knowledge graph,'' \emph{Pattern Recognition Letters}, vol. 143, pp. 43--49,
  2021.

\bibitem{li2024learning}
Y.~Li, X.~Zhang, X.~Cheng, X.~Tang, and L.~Jiao, ``Learning consensus-aware
  semantic knowledge for remote sensing image captioning,'' \emph{Pattern
  Recognition}, vol. 145, p. 109893, 2024.

\bibitem{das2024textgcn}
S.~Das and R.~Sharma, ``A textgcn-based decoding approach for improving remote
  sensing image captioning,'' \emph{IEEE Geoscience and Remote Sensing
  Letters}, vol.~22, pp. 1--5, 2024.

\bibitem{liu2025semantic}
M.~Liu, J.~Liu, and X.~Zhang, ``Semantic-spatial feature fusion with dynamic
  graph refinement for remote sensing image captioning,'' \emph{IEEE Journal of
  Selected Topics in Applied Earth Observations and Remote Sensing}, 2025.

\bibitem{sun2025scene}
Q.~Sun, Y.~Wang, and X.~Song, ``Scene graph and dependency grammar enhanced
  remote sensing change caption network (sgd-rsccn),'' in \emph{Proceedings of
  the 31st International Conference on Computational Linguistics}, 2025, pp.
  2121--2130.

\bibitem{li2026knowledge}
G.~Li, D.~Hu, H.~Li, Z.~Yao, W.~Mi, Z.~Liu, X.~Zhang, and H.~Lyu,
  ``Knowledge-enhanced image captioning with adaptive graph-based multimodal
  alignment and llm,'' in \emph{Proceedings of the AAAI Conference on
  Artificial Intelligence}, vol.~40, no.~18, 2026, pp. 15\,090--15\,098.

\bibitem{park2019robust}
D.~H. Park, T.~Darrell, and A.~Rohrbach, ``Robust change captioning,'' in
  \emph{Proceedings of the IEEE/CVF International Conference on Computer
  Vision}, 2019, pp. 4624--4633.

\bibitem{qiu2021describing}
Y.~Qiu, S.~Yamamoto, K.~Nakashima, R.~Suzuki, K.~Iwata, H.~Kataoka, and
  Y.~Satoh, ``Describing and localizing multiple changes with transformers,''
  in \emph{Proceedings of the IEEE/CVF International Conference on Computer
  Vision}, 2021, pp. 1971--1980.

\bibitem{zhou2024single}
Q.~Zhou, J.~Gao, Y.~Yuan, and Q.~Wang, ``Single-stream extractor network with
  contrastive pre-training for remote-sensing change captioning,'' \emph{IEEE
  Transactions on Geoscience and Remote Sensing}, vol.~62, pp. 1--14, 2024.

\bibitem{yu2025diffusion}
X.~Yu, Y.~Li, J.~Ma, C.~Li, and H.~Wu, ``Diffusion-rscc: Diffusion
  probabilistic model for change captioning in remote sensing images,''
  \emph{IEEE Transactions on Geoscience and Remote Sensing}, 2025.

\bibitem{wang2024ringmogpt}
P.~Wang, H.~Hu, B.~Tong, Z.~Zhang, F.~Yao, Y.~Feng, Z.~Zhu, H.~Chang, W.~Diao,
  Q.~Ye \emph{et~al.}, ``Ringmogpt: A unified remote sensing foundation model
  for vision, language, and grounded tasks,'' \emph{IEEE Transactions on
  Geoscience and Remote Sensing}, vol.~63, pp. 1--20, 2024.

\bibitem{wang2026kgbdcnet}
D.~Wang, G.~Ma, X.~Wang, Y.~Zhang, H.~Zhang, B.~Wang, and P.~Chen, ``Kgbdcnet:
  Keyword-guided building damage captioning network for bi-temporal remote
  sensing images,'' \emph{ISPRS Journal of Photogrammetry and Remote Sensing},
  vol. 234, pp. 369--385, 2026.

\bibitem{xian2026cross}
T.~Xian, Z.~Zhou, W.~Zhou, D.~Zeng, and B.~Li, ``Cross-view and multi-step
  interaction for change captioning,'' \emph{IEEE Transactions on Multimedia},
  2026.

\bibitem{deng2026deltavlm}
P.~Deng, W.~Zhou, and H.~Wu, ``Deltavlm: Interactive remote sensing image
  change analysis via instruction-guided difference perception,'' \emph{Remote
  Sensing}, vol.~18, no.~4, p. 541, 2026.

\bibitem{liu2023decoupling}
C.~Liu, R.~Zhao, J.~Chen, Z.~Qi, Z.~Zou, and Z.~Shi, ``A decoupling paradigm
  with prompt learning for remote sensing image change captioning,'' \emph{IEEE
  Transactions on Geoscience and Remote Sensing}, vol.~61, pp. 1--18, 2023.

\bibitem{zhu2025change3d}
D.~Zhu, X.~Huang, H.~Huang, H.~Zhou, and Z.~Shao, ``Change3d: Revisiting change
  detection and captioning from a video modeling perspective,'' in
  \emph{Proceedings of the Computer Vision and Pattern Recognition Conference},
  2025, pp. 24\,011--24\,022.

\bibitem{wang2026dfm}
Y.~Wang, Z.~Song, C.~Yang, M.~Wang, Z.~An, L.~Huang, and Y.~Xu, ``Dfm:
  Difference feature modeling with text-guided gated contrastive loss for
  remote sensing image change captioning,'' \emph{arXiv preprint
  arXiv:2606.27410}, 2026.

\bibitem{wang2026sam}
F.~Wang, M.~Wang, X.~Wang, H.~Wang, and J.~Tang, ``Sam-guided semantic and
  motion changed region mining for remote sensing change captioning,''
  \emph{IEEE Transactions on Geoscience and Remote Sensing}, 2026.

\bibitem{chen2020spatial}
H.~Chen and Z.~Shi, ``A spatial-temporal attention-based method and a new
  dataset for remote sensing image change detection,'' \emph{Remote sensing},
  vol.~12, no.~10, p. 1662, 2020.

\bibitem{papineni2002bleu}
K.~Papineni, S.~Roukos, T.~Ward, and W.-J. Zhu, ``Bleu: a method for automatic
  evaluation of machine translation,'' in \emph{Proceedings of the 40th annual
  meeting of the Association for Computational Linguistics}, 2002, pp.
  311--318.

\bibitem{banerjee2005meteor}
S.~Banerjee and A.~Lavie, ``Meteor: An automatic metric for mt evaluation with
  improved correlation with human judgments,'' in \emph{Proceedings of the acl
  workshop on intrinsic and extrinsic evaluation measures for machine
  translation and/or summarization}, 2005, pp. 65--72.

\bibitem{lin2004rouge}
C.-Y. Lin, ``Rouge: A package for automatic evaluation of summaries,'' in
  \emph{Text summarization branches out}, 2004, pp. 74--81.

\bibitem{vedantam2015cider}
R.~Vedantam, C.~Lawrence~Zitnick, and D.~Parikh, ``Cider: Consensus-based image
  description evaluation,'' in \emph{Proceedings of the IEEE conference on
  computer vision and pattern recognition}, 2015, pp. 4566--4575.

\bibitem{paszke2019pytorch}
A.~Paszke, S.~Gross, F.~Massa, A.~Lerer, J.~Bradbury, G.~Chanan, T.~Killeen,
  Z.~Lin, N.~Gimelshein, L.~Antiga \emph{et~al.}, ``Pytorch: An imperative
  style, high-performance deep learning library,'' \emph{Advances in neural
  information processing systems}, vol.~32, 2019.

\bibitem{sun2025mask}
D.~Sun, J.~Yao, W.~Xue, C.~Zhou, P.~Ghamisi, and X.~Cao, ``Mask approximation
  net: A novel diffusion model approach for remote sensing change captioning,''
  \emph{IEEE Transactions on Geoscience and Remote Sensing}, 2025.

\bibitem{bai2025cross}
Q.~Bai and X.~Wang, ``Cross-temporal remote sensing image change captioning: A
  manifold mapping and bayesian diffusion approach for land use monitoring,''
  \emph{IEEE Journal of Selected Topics in Applied Earth Observations and
  Remote Sensing}, 2025.

\bibitem{gao2026uav}
Y.~Gao, T.~Li, G.~Wang, and Y.~Yang, ``Uav as urban construction change
  monitor: A new benchmark and change captioning model,'' \emph{arXiv preprint
  arXiv:2605.04409}, 2026.

\bibitem{peng2026frequency}
C.~Peng, F.~Wu, W.~Song, and Z.~Wang, ``Frequency-spatial semantic decoupling
  for remote sensing image change captioning,'' \emph{IEEE Journal of Selected
  Topics in Applied Earth Observations and Remote Sensing}, 2026.

\end{thebibliography}
}

\end{document}